\documentclass[journal]{IEEEtran}

\ifCLASSINFOpdf
\else
   \usepackage[dvips]{graphicx}
\fi
\usepackage{url}
\usepackage{enumitem}
\usepackage{bm}
\usepackage{amsmath}
\usepackage{amssymb}
\newcommand{\given}{\,|\,}

\usepackage{booktabs}
\usepackage{multirow}
\usepackage{makecell}
\usepackage{xcolor}
\usepackage{orcidlink}
\usepackage{balance}

\usepackage{graphicx}

\begin{document}

\title{Linearized Coupling Flow with Shortcut Constraints for One-Step Face Restoration}

\author{Xiaohui Sun, Hanlin Wu\,\orcidlink{0000-0002-3505-0521}, \IEEEmembership{Member, IEEE}
    \thanks{This work was supported in part by the National Natural Science Foundation of China under Grant 62401064, and in part by the Fundamental Research Funds for the Central Universities under Grant 2024JJ040 and Grant 2024TD001. \emph{(Corresponding author: Hanlin Wu.)}
    }
    \thanks{The authors are with the School of Information Science and Technology, Beijing Foreign Studies University, Beijing 100089, China (e-mail: hlwu@bfsu.edu.cn).}
}

\maketitle

\begin{abstract}
Face restoration can be formulated as a continuous-time transformation between image distributions via Flow Matching (FM). However, standard FM typically employs independent coupling, ignoring the statistical correlation between low-quality (LQ) and high-quality (HQ) data. This leads to intersecting trajectories and high velocity-field curvature, requiring multi-step integration. We propose Shortcut-constrained Coupling Flow for Face Restoration (SCFlowFR) to address these challenges. By establishing a data-dependent coupling, we explicitly model the LQ--HQ dependency to minimize path crossovers and promote near-linear probability flow. Furthermore, we employ a conditional mean estimator to refine the source distribution’s anchor, effectively tightening the transport cost and stabilizing the velocity field. To ensure stable one-step inference, a shortcut constraint is introduced to supervise average velocities over arbitrary intervals, mitigating discretization bias in large-step updates. SCFlowFR achieves state-of-the-art one-step restoration, providing a superior trade-off between perceptual fidelity and computational efficiency.
\end{abstract}


\IEEEpeerreviewmaketitle

\section{Introduction}

\IEEEPARstart{F}{ace} restoration, a classical ill-posed inverse problem, seeks to reconstruct high-quality (HQ) facial images from corrupted low-quality (LQ) observations. It is fundamental to various signal processing applications, including biometric identification and video communication~\cite{wang2025freqformer,wu2025video}. While recent generative paradigms have significantly advanced restoration fidelity, a critical trade-off persists between reconstruction quality and efficiency~\cite{luo2023latent,yin2024one}.

Diffusion Models (DMs) formulate restoration as a gradual refinement process~\cite{yue2023resshift,wu2024seesr,wang2024exploiting}, achieving impressive perceptual quality and fidelity by integrating a stochastic differential equation (SDE) \cite{song2020score}. However, their iterative nature requires dozens of sampling steps, leading to high inference latency that hinders real-time signal processing. Recently, Flow Matching (FM) \cite{lipman2023flow, liu2023rectified} has emerged as an efficient deterministic alternative. FM learns a time-dependent velocity field $\bm{v}_{\theta}$ that satisfies the continuity equation, defining an ordinary differential equation (ODE) to transport a source distribution $\rho_0$ to the target HQ distribution $\rho_1$.

\begin{figure}[t]
    \centering
    \centerline{\includegraphics[width=\linewidth]{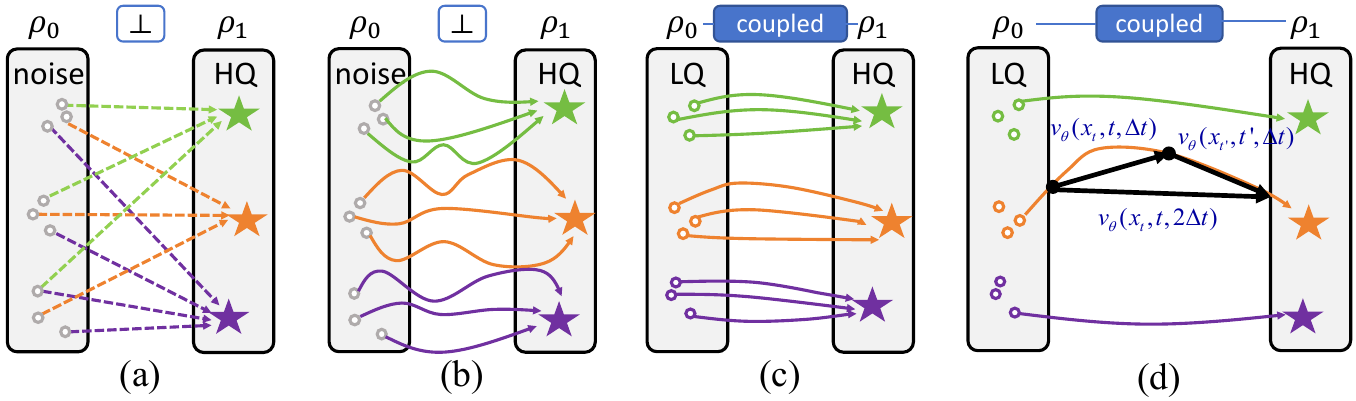}}
    \caption{(a) Intersecting paths in independent coupling. (b) Resulting curved velocity field in vanilla FM. (c) Straighter trajectories via data-dependent coupling. (d) One-step sampling via shortcut-constrained average velocity.}
    \label{fig:intro}
    \vspace{-0.5em}
\end{figure}

Despite its potential, standard FM-based restoration~\cite{zhu2024flowie, martin2024pnp} typically assumes an \textit{independent coupling}, where source samples (e.g., Gaussian noise) and target HQ images are paired without considering the observed LQ signal. From a statistical perspective, this ignores the inherent conditional dependency between the LQ input and the HQ ground truth. As illustrated in Fig.\,\ref{fig:intro}\,(a), independent pairing yields intersecting interpolation paths. Since the resulting ODE flow is deterministic and non-intersecting, the model is forced to learn a highly non-linear velocity field with significant trajectory curvature (Fig.\,\ref{fig:intro}\,(b)). In such curved dynamics, the approximation of the instantaneous derivative during one-step inference is highly susceptible to discretization errors, leading to degraded fidelity~\cite{frans2024one,cohen2025efficient}.

\begin{figure*}[t]
    \centering
    \centerline{\includegraphics[width=0.95\linewidth]{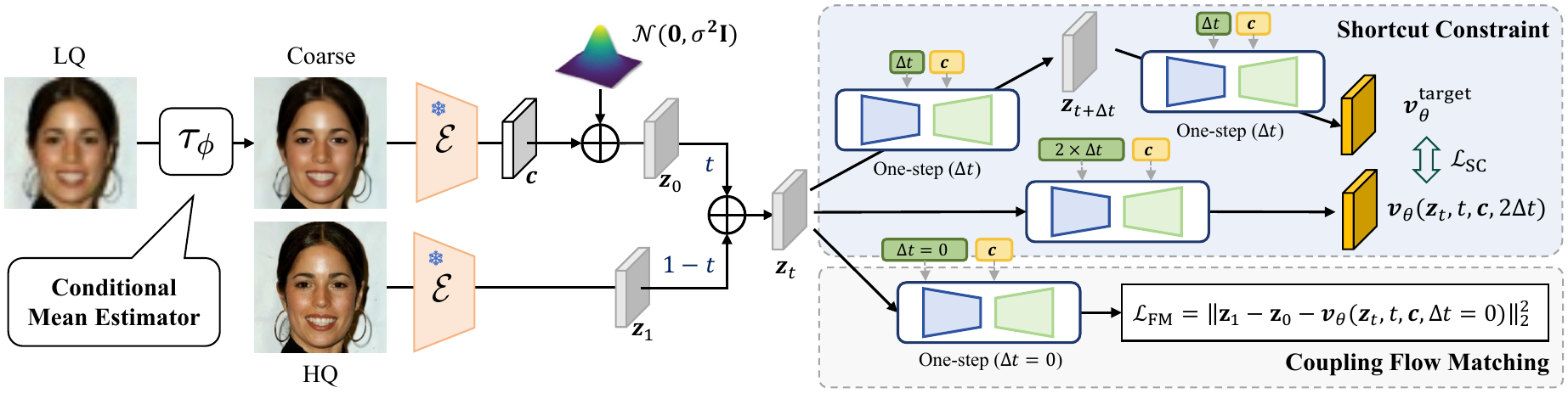}}
    \caption{Overview of SCFlowFR. We construct a coupled LQ--HQ transport path (left), conditioning the velocity field $\bm{v_{\theta}}$ on a coarse estimate $\bm{c}$ from $\tau_\phi$. A shortcut constraint then enables average velocity prediction over time interval $\Delta t$.}
    \label{fig:method}
    \vspace{-0.5em}
\end{figure*}

To address the challenges of non-linear transport and integration instability, we propose Shortcut-constrained Coupling Flow for Face Restoration (SCFlowFR). Our framework optimizes the transport dynamics through three key innovations. First, we establish a \textit{data-dependent coupling} that reformulates the transport as a conditional process. By leveraging the mutual information between the LQ and HQ domains, we minimize path crossovers and promote near-linear probability flow (Fig.\,\ref{fig:intro}\,(c)). Second, we utilize a \textit{conditional mean estimation} as a Bayesian-like prior to anchor the source distribution. This strategy effectively minimizes the expected squared displacement (transport cost) and stabilizes the learned velocity field. Third, we integrate a \textit{shortcut constraint} that supervises the model to predict the average velocity over arbitrary time intervals rather than the instantaneous derivative. This enables the model to implicitly compensate for residual curvature, ensuring stable and accurate one-step inference even under complex degradations (Fig.\,\ref{fig:intro}\,(d)).

The main contributions can be summarized as follows:
\begin{enumerate}[label=\arabic*),itemsep=0pt,topsep=0pt]
    \item We propose a novel FM framework that establishes a \textit{data-dependent coupling}, explicitly modeling the LQ--HQ dependency to linearize the transport trajectory.
    \item We employ a \textit{conditional mean estimator} that refines the source distribution's center, thereby tightening the transport cost bound and reducing velocity field variance.
    \item We introduce a \textit{shortcut constraint} that supervises average velocities over arbitrary time intervals, enabling stable and accurate one-step inference.
    \item Extensive experiments demonstrate that our method achieves state-of-the-art (SOTA) restoration quality with efficiency comparable to non-iterative baselines.
\end{enumerate}

\section{Methodology}
This section details SCFlowFR framework for efficient one-step face restoration, as illustrated in Fig.\,\ref{fig:method}.

\subsection{Background on FM}

FM \cite{lipman2023flow} transforms samples from a source distribution $\rho_0$ to a target $\rho_1$ by learning a parameterized velocity field $\bm{v}_\theta(\bm{x}, t)$ that defines a probability flow ordinary differential equation (ODE):
\begin{equation}
    \dot{\bm{x}}_t = \bm{v}_\theta(\bm{x}_t, t), \quad t \in [0,1].
\end{equation}
In standard FM, the source and target are typically assumed independent, i.e., $\rho(\bm{x}_0, \bm{x}_1) = \rho_0(\bm{x}_0)\rho_1(\bm{x}_1)$. Training uses straight-line interpolants $(1-t)\bm{x}_0 + t\bm{x}_1$ to define target velocities. However, since $\bm{x}_0$ and $\bm{x}_1$ are arbitrarily paired, their interpolated paths frequently cross at intermediate times, as illustrated in Fig.\,\ref{fig:intro}\,(a). Since ODE flows are deterministic and non-intersecting, the model must warp these paths into complex, winding trajectories---forcing $\bm{v}_\theta$ to be highly nonlinear, as illustrated in Fig.\,\ref{fig:intro}\,(b), which harms both learning accuracy and integration stability.

\subsection{Data-Dependent Coupling for Linearized Transport}

We establish a data-dependent coupling between the source and target distributions, tailored for the inherent pairing in face restoration. Since LQ and HQ images share substantial semantic and structural content, we can naturally condition the source distribution on the target. 

Following recent practices~\cite{zhu2024flowie,lin2024diffbir}, we model this coupling in a compact latent space. Let $\mathcal{E}(\cdot)$ denote a pretrained encoder. We define the HQ target latent as $\bm{z}_1=\mathcal{E}(\mathrm{HQ}) \sim \rho_1(\bm{z}_1)$. For each $\bm{z}_1$, we construct a conditional source latent $\bm{z}_0$ by encoding the corresponding LQ input, defining the joint coupling as:
\begin{equation}
\rho(\bm{z}_0, \bm{z}_1) = \rho_1(\bm{z}_1) \, \rho(\bm{z}_0 \given \bm{z}_1),
\end{equation}
where $\rho(\bm{z}_0 \given \bm{z}_1)$ is a narrow distribution centered at the encoded degraded image. Specifically,
\begin{equation}
    \bm{z}_0 = \mathcal{E}(\mathrm{LQ}) + \bm\varepsilon, \quad \bm\varepsilon\sim \mathcal{N}(0, \sigma^2 \mathbf{I}),
\end{equation}
with a small variance $\sigma\in[0.03, 0.08]$ ensuring smooth density while preserving tight coupling. This formulation defines the straight-line interpolation $\bm{z}_t = (1 - t) \bm{z}_0 + t \bm{z}_1$ and yields two key properties (see Appendix~A for a formal analysis):

\textbf{Property 1: Reduced Conditional Velocity Variance.}
The ODE is driven by the marginal velocity $\bm{v}_t(\bm{z})$, defined as the conditional expectation of individual linear paths~\cite{albergo2024stochastic}:
\begin{equation}
\label{eq:velocity_expectation}
\bm{v}_t(\bm{z}) = \mathbb{E}[\bm{z}_1 - \bm{z}_0 \mid \bm{z}_t = \bm{z}].
\end{equation}

Under independent coupling, multiple distinct pairs $(\bm{z}_0, \bm{z}_1)$ can correspond to the same intermediate state $\bm{z}_t$, leading to high conditional variance of $\bm{z}_1 - \bm{z}_0$ and thus an ambiguous velocity field. Conversely, our tight conditional coupling $\rho(\bm{z}_0\mid\bm{z}_1)$ makes the pair nearly deterministic given $\bm{z}_t$, reducing this variance and straightening the trajectories.

\textbf{Property 2: Reduced Transport Kinetic Energy.} 
For the ODE solution $\bm{Z}_t$, the expected kinetic energy is upper bounded by the expected squared displacement between endpoints:
\begin{equation}
\label{eq:upper_bound}
\mathbb{E} \int_0^1 \|\dot{\bm{Z}}_t\|^2 \, dt \leq \mathbb{E}[\|\bm{z}_1 - \bm{z}_0\|^2].
\end{equation}
Independent pairing inflates this bound, leading to inefficient transport. By anchoring $\bm{z}_0$ near the semantically aligned $\bm{z}_1$, our coupling effectively minimizes $\mathbb{E}[\|\bm{z}_1 - \bm{z}_0\|^2]$. This aligns the dynamics closer to the Optimal Transport (OT) plan, mitigating discretization errors crucial for stable few-step inference.

\subsection{Conditional Mean Estimation for Tight Coupling}
\label{subsec:condition_mean_estimation}

While data-dependent coupling structurally improves the flow, the tightness of the bound in Eq.\,\eqref{eq:upper_bound} depends critically on the alignment between $\bm{z}_0$ and $\bm{z}_1$. Under severe real-world degradations, the raw LQ observation often lacks reliable structural cues, causing its encoded latent $\mathcal{E}(\mathrm{LQ})$ to deviate significantly from $\bm{z}_1$, which inflates the transport cost.

To mitigate this, we observe that $\mathbb{E}[\|\bm{z}_1 - \bm{z}_0\|^2]$ is minimized when $\bm{z}_0$ is centered at the conditional mean $\mathbb{E}[\bm{z}_1 \given \mathrm{LQ}]$. Although this mean is intractable, a regression-based restorer $\tau_\phi$, trained with $\ell_2$ loss on synthetic data, provides a practical approximation:
\begin{equation}
    \bm{c} = \mathcal{E}\bigl(\tau_\phi(\mathrm{LQ})\bigr) \approx \mathbb{E}[\mathcal{E}(\mathrm{HQ}) \mid \mathrm{LQ}],
\end{equation}
which serves as a more accurate center for the source distribution. We redefine the conditional source distribution as
\begin{equation}
    \rho(\bm{z}_0 \given \bm{z}_1) = \mathcal{N}(\bm{c}, \sigma^2 \mathbf{I}),
\end{equation}
further tightening the upper bound in Eq.\,(\ref{eq:upper_bound}), yielding straighter trajectories and a smoother velocity field.

Furthermore, $\bm{c}$ acts as a semantic conditioning signal for the velocity network $\bm{v}_\theta(\bm{z}_t, t, \bm{c})$, providing crucial guidance for accurate velocity estimation. Consequently, we train $\bm{v}_\theta$ to approximate the target velocity field in Eq.\,\eqref{eq:velocity_expectation} by regressing the ground-truth displacement $\bm{z}_1 - \bm{z}_0$:
\begin{equation}
\label{eq:fm_loss}
\mathcal{L}_{\text{FM}}(\theta)
= \mathbb{E}_{\bm{z}_0,\bm{z}_1,t}
\bigl\|\bm{v}_\theta(\bm{z}_t,t,\bm{c})-(\bm{z}_1-\bm{z}_0)\bigr\|^2.
\end{equation}

\begin{table*}[t]
\centering
\setlength{\tabcolsep}{6pt}
\renewcommand{\arraystretch}{0.9}
\caption{Quantitative results on the CelebA-Test dataset. Among all one-step methods, best in \textcolor{red}{red}, second-best in \textcolor{blue}{blue}.}
\label{tab:celeba0}
\begin{tabular}{l l c c c c c c c c}
\toprule[1.5pt]
\multicolumn{1}{l}{\multirow{2}{*}{Type}} & \multicolumn{1}{l}{\multirow{2}{*}{Method}} & \multicolumn{1}{c}{\multirow{2}{*}{Steps}} & \multicolumn{2}{c}{Efficiency} & \multicolumn{5}{c}{Quality Metrics} \\
\cmidrule(lr){4-5}\cmidrule(lr){6-10}
& & & Params (M)\,$\downarrow$ & FPS\,$\uparrow$
& FID\,$\downarrow$ & PSNR (dB)\,$\uparrow$ & LPIPS\,$\downarrow$ & MUSIQ\,$\uparrow$ & BRISQUE\,$\downarrow$ \\
\midrule

\multirow{4}{*}{\makecell[c]{Multi-step}}
& DiffBIR \cite{lin2024diffbir} & 50 & 1667 & 0.11 & 16.91 & 24.15 & 0.30 & 76.05 & -1.09  \\
& StableSR \cite{wang2024exploiting} & 30 & 1410 & 5.80 & 16.96 & 24.33 & 0.27 & 71.56 & 20.64 \\
& PMRF \cite{ohayon2024posterior} & 25 & 183 & 2.85 & 21.70 & 23.59 & 0.31 & 61.79 & 24.49  \\
& FlowIE \cite{zhu2024flowie} & 5 & 1717 & 1.17 & 17.10 & 23.45 & 0.27 & 73.38 & 2.28  \\
\midrule

\multirow{5}{*}{\makecell[c]{One-step}}
& DMDNet \cite{li2022learning} & 1 & \textcolor{blue}{40} & 8.40 & 29.85 & 24.09 & 0.30 & 69.76 & 19.26  \\
& RestoreFormer \cite{wang2022restoreformer} & 1 & \textcolor{red}{17} & \textcolor{blue}{22.96} & 21.22 & 23.27 & 0.30 & \textcolor{blue}{72.55} &  \textcolor{red}{-4.69}  \\
& OSEDiff \cite{wu2024one} & 1 & 1775 & 3.91 & 24.27 & 23.12 &  \textcolor{red}{0.27} & 70.23 & 18.95  \\
& SCFlowFR-Tiny (Ours) & 1 & 96 & \textcolor{red}{34.32} &  \textcolor{blue}{16.36} & \textcolor{blue}{24.25} & 0.31 & 69.71 & 0.90  \\
& SCFlowFR (Ours) & 1 & 405 & 8.62 & \textcolor{red}{15.62} & \textcolor{red}{24.26}& \textcolor{blue}{0.29} & \textcolor{red}{72.66} & \textcolor{blue}{-1.54}  \\

\bottomrule[1.5pt]
\end{tabular}
\end{table*}

\subsection{Shortcut Constraints for One-Step Inference}

Although our conditional coupling produces near-linear trajectories with reduced transport cost, one-step integration still incurs endpoint errors due to imperfect velocity estimation and coarse discretization. To mitigate this, we introduce a shortcut constraint~\cite{frans2024one} that corrects the bias by learning the average velocity, enabling high-fidelity one-step inference.

To facilitate accurate large-step integration, we equip the velocity network with step-size awareness. Instead of predicting only the instantaneous derivative, $\bm{v}_\theta$ learns to output the \textit{average velocity} over an arbitrary time interval $\Delta t$, defined as:
\begin{equation}
\label{eq:euler_step}
\bm{z}_{t+\Delta t}' = \bm{z}_t + \bm{v}_\theta(\bm{z}_t, t, \bm{c}, \Delta t)\cdot\Delta t.
\end{equation}
As $\Delta t \to 0$, this shortcut direction naturally reduces to the instantaneous velocity field (Eq.\,\eqref{eq:velocity_expectation}); for $\Delta t > 0$, the model predicts the integrated motion over larger increments.

We achieve this by leveraging a self-consistency constraint, ensuring that taking one large shortcut step of size $2\Delta t$ from state $\bm{z}_t$ yields the same displacement as taking two consecutive steps of size $\Delta t$. To stabilize training and avoid collapse, target velocities are computed via an Exponential Moving Average (EMA) network parameterized by $\theta^-$:
\begin{equation}
    \bm{v}_{\theta^-}^{\text{target}} = \frac{1}{2} \big[\bm{v}_{\theta^-}(\bm{z}_t, t, \bm{c}, \Delta t)
    + \bm{v}_{\theta^-}(\bm{z}'_{t+\Delta t}, t+\Delta t, \bm{c}, \Delta t) \big].
\end{equation}
The corresponding shortcut training objective $\mathcal{L}_{\text{SC}}$ is given by:
\begin{equation}
\label{eq:sc_loss}
\mathcal{L}_{\text{SC}}(\theta) = \mathbb{E}_{\bm{z}_0,\bm{z}_1,t,\Delta t}
{\bigl\|\bm{v}_\theta(\bm{z}_t, t, \bm{c}, 2\Delta t) - \bm{v}_{\theta^-}^{\text{target}}\bigr\|^2}.
\end{equation}

During training, each batch is divided into two subsets: one optimizes the FM loss (Eq.\,\ref{eq:fm_loss}) with $\Delta t = 0$, and the other enforces the shortcut constraint (Eq.\,\ref{eq:sc_loss}) with randomly sampled $\Delta t > 0$, jointly learning an LQ-to-HQ mapping with consistent dynamics across step sizes.

During inference, given an LQ image, we compute the conditional prior $\bm{c} = \mathcal{E}\bigl(\tau_\phi(\mathrm{LQ})\bigr)$, sample the anchored source $\bm{z}_0 \sim \mathcal{N}(\bm{c}, \sigma^2 \mathbf{I})$, and generate the target latent in a single step:
\begin{equation}
    \hat{\bm{z}}_1 = \bm{z}_0 + \bm{v}_\theta(\bm{z}_0, 0, \bm{c}, \Delta t=1).
\end{equation}
Finally, $\hat{\bm{z}}_1$ is passed through a pretrained decoder $\mathcal{D}$ to reconstruct the restored HQ image.

\section{Experiments}
\subsection{Experiment Setups}
\label{ssec:experiment_setup}
\textbf{Datasets.} We train SCFlowFR on the FFHQ dataset \cite{karras2019style}, which contains 70,000 HR images. All images are resized to 512$\times$512 resolution before training. During evaluation, we first assess SCFlowFR performance on the synthetic dataset CelebA-Test \cite{liu2015deep}, which contains 3,000 paired HQ and LQ images. To further validate the generalization capability of SCFlowFR in real-world scenarios, we conduct additional experiments on three wild datasets: LFW-Test \cite{wang2021towards}, CelebChild-Test \cite{wang2021towards}, and WebPhoto-Test \cite{wang2021towards}, all of which comprise facial images affected by varying degrees of degradation.

\begin{figure*}[t]

  \centering
  \includegraphics[width=0.9\linewidth]{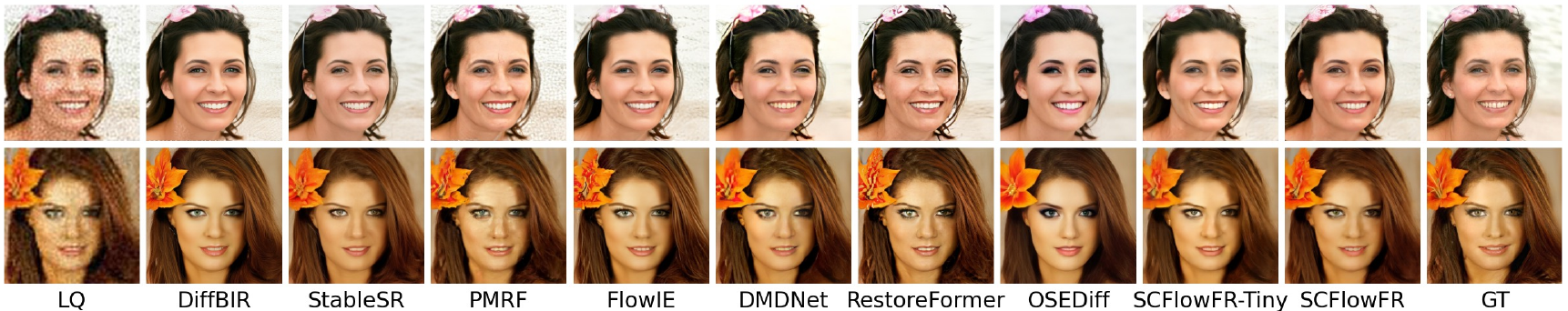}
  \caption{Qualitative comparisons on the CelebA-Test dataset.}
  \label{fig:experiment_celeb}
  \vspace{-1em}
\end{figure*}

\textbf{Training Details.}  For both training and evaluation, we used the synthetic degradation pipeline of Codeformer \cite{wang2021towards}. A pre-trained SwinIR \cite{liang2021swinir} served as the conditional mean estimator $\tau_\phi$, while $\bm{v}_{\theta}$ was implemented as a U-Net \cite{ronneberger2015u}. The parameters of SwinIR and the VAE were frozen. Training was performed with a batch size of 16 for 200k iterations using the Adam optimizer and a learning rate $1\times 10^{-4}$. Moreover, we trained a lightweight variant, SCFlowFR-Tiny, by replacing the VAE and U-Net with compact architectures \cite{taesd}, training it for 150k iterations under the same settings.

\textbf{Metrics.} On synthetic datasets, we employed full-reference image accessment metrics including FID, PSNR, and LPIPS and no-reference metrics, MUSIQ \cite{ke2021musiq} and BRISQUE \cite{mittal2012no}. For wild datasets, evaluation was conducted exclusively using the no-reference metrics NIQE and BRISQUE. Efficiency was assessed via comparisons of parameter counts and inference throughput (FPS).

\subsection{Comparisons with State-of-the-Art Methods}
\label{ssec:subhead}
We compare our method with the following baselines: DMDNet \cite{li2022learning}, RestoreFormer \cite{wang2022restoreformer}, StableSR \cite{wang2024exploiting}, DiffBIR \cite{lin2024diffbir}, OSEDiff \cite{wu2024one}, PMRF \cite{ohayon2024posterior}, and FlowIE \cite{zhu2024flowie}. For comparison, we categorize them by inference steps: DMDNet, RestoreFormer, and OSEDiff are one-step methods, while StableSR, DiffBIR, PMRF, and FlowIE are multi-step methods.

\textbf{CelebA-Test.} 
As shown in Table \ref{tab:celeba0}, among one-step methods, SCFlowFR achieves the best scores on FID, PSNR, and MUSIQ, while securing second-best on the remaining metrics, demonstrating its comprehensive restoration quality. The lightweight variant SCFlowFR-Tiny also ranks highly, confirming that a compact architecture retains strong perceptual performance. More importantly, compared to computationally expensive multi-step approaches, our method establishes a superior efficiency-quality trade-off: it delivers top-tier restoration in a single step with runtime orders of magnitude faster than iterative diffusion or FM methods, enabling real-time applications.
We further present qualitative results in Fig.\,\ref{fig:experiment_celeb}. SCFlowFR effectively preserves crucial image information, avoiding the over-generation or distortion caused by noise interference, while successfully restoring fine-grained details. Notably, it achieves a level of visual fidelity comparable to that of multi-step inference models, resulting in realistic and highly faithful restoration.

\textbf{Wild Datasets.}
We conducted comparative experiments using the four-step inference results from both SCFlowFR and its lightweight variant, benchmarking against relevant SOTA methods. As shown in Table \ref{tab:nr_metrics_comparison},  SCFlowFR-Tiny achieves superior NIQE and BRISQUE scores across nearly all three wild datasets. Its advantage over the larger variant stems from better alignment with mild, less structured real-world degradations, yielding more natural restoration without over-parameterization. Further qualitative results are provided in Appendix~B.

\begin{table}[t]
\centering
\setlength{\tabcolsep}{5pt}
\caption{No-reference metrics on three wild datasets. Best in \textcolor{red}{red}, second-best in \textcolor{blue}{blue}. ``BR.'' denotes BRISQUE.}
\label{tab:nr_metrics_comparison}
\begin{tabular}{@{}lcccccc@{}}
\toprule[1.5pt]
\multirow{2}{*}{Method} & \multicolumn{2}{c}{LFW} & \multicolumn{2}{c}{CelebChild} & \multicolumn{2}{c}{WebPhoto} \\
\cmidrule(lr){2-3} \cmidrule(lr){4-5} \cmidrule(lr){6-7}
 & NIQE\,$\downarrow$ & BR.\,$\downarrow$ & NIQE\,$\downarrow$ & BR.\,$\downarrow$ & NIQE\,$\downarrow$ & BR.\,$\downarrow$ \\
\midrule
StableSR \cite{wang2024exploiting} & 5.24 & -0.22 & 4.91 & 3.23 & 5.78 & \textcolor{blue}{2.02} \\
DiffBIR \cite{lin2024diffbir} & 5.68 & \textcolor{red}{-3.09} & 5.48 & \textcolor{blue}{1.40} & 6.01 & 4.58 \\
OSEDiff \cite{wu2024one} & 4.72 & 2.78 & 5.07 & 9.82 & 5.26 & 5.89 \\
FlowIE \cite{zhu2024flowie} & 4.27 & 1.16 & 4.44 & 7.39 & 4.80 & 6.83 \\
SCFlowFR & \textcolor{blue}{4.11} & \textcolor{blue}{-1.01} & \textcolor{blue}{4.25} & 3.23 & \textcolor{blue}{4.38} & 2.89 \\
SCFlowFR-Tiny & \textcolor{red}{3.96} & 1.65 & \textcolor{red}{4.04} & \textcolor{red}{0.82} & \textcolor{red}{4.23} & \textcolor{red}{1.97} \\
\bottomrule[1.5pt]
\end{tabular}
\vspace{-0.5em}
\end{table}

\subsection{Ablation Studies}
\label{ssec:ablation}

\textbf{Discussion on Training Strategies.}
To validate the necessity of the shortcut training strategy and investigate the impact of different training strategies on performance, we designed corresponding ablation studies. 
First, we compared a model trained solely with the $\mathcal{L}_{\text{FM}}$ loss, denoted as Model-w/o-$\mathcal{L}_{\text{SC}}$. As shown in Table 3, the results demonstrate significantly degraded performance. Second, we evaluated the training strategy proposed by Consistency-FM \cite{yang2024consistency}, denoted as Model-w/-$\mathcal{L}_{\text{consis}}$. Since the model failed to achieve satisfactory performance under one-step inference, we report its five-step results, which still indicate inadequate performance.

\textbf{Discussion on Conditional Guidance.} Our conditional guidance strategy leverages a pre-restored image as the contextual input to the velocity network, providing a semantically aligned prior for the HQ target. To verify its contribution, we train an ablated model that removes this conditional branch, denoted Model-w/o-cond. As shown in Table~\ref{tab:ablation}, the absence of conditional guidance leads to consistent degradation in FID, PSNR, and BRISQUE, confirming that the pre-restored image as a conditional signal is essential for achieving accurate and robust one-step restoration.
\begin{table}[t]
\centering
\setlength{\tabcolsep}{7pt}
\caption{Ablation studies on the CelebA-Test dataset. Best performance is highlighted in \textcolor{red}{red}.}
\label{tab:ablation}
\begin{tabular}{lcccc}
\toprule[1.5pt]
Method & Steps & FID\,$\downarrow$ & PSNR(dB)\,$\uparrow$ & BRISQUE\,$\downarrow$ \\
\midrule
Model-w/o-$\mathcal{L}_{\text{SC}}$ & 1 & 287.23 & 20.85 & 51.22 \\
Model-w/-$\mathcal{L}_{\text{consis}}$ & 5 & 18.54 & 24.04 & 3.17 \\
Model-w/o-cond & 1 & 19.01 & 23.08 & 1.55 \\
Ours & 1 & \color{red}{16.36} & \color{red}{24.25} & \color{red}{0.90} \\
\bottomrule[1.5pt]
\end{tabular}
\vspace{-0.5em}
\end{table}
\section{Conclusion}
\label{sec:conclusion}
We propose SCFlowFR, a face restoration method based on Shortcut-constrained coupling FM. Unlike conventional approaches starting from Gaussian noise, SCFlowFR establishes data-dependent coupling by centering the source on a coarse reconstruction from the LQ input and uses it for conditional guidance---reducing path crossovers and promoting near-linear transport. A shortcut constraint enforces self-consistency via average velocity supervision, enabling accurate one-step inference. Experiments show SCFlowFR achieves SOTA one-step quality with efficiency comparable to non-iterative baselines. Future work will extend this framework to general image restoration and inpainting to enhance robustness and versatility.

\balance

\bibliographystyle{IEEEtran}
\bibliography{references}

\newpage
\begin{center}
{\small\bf SUPPLEMENTARY MATERIAL}
\end{center}

Appendix A contains mathematical proofs for the two properties in Section II, demonstrating how our data-dependent coupling mitigates path crossover and curvature in standard FM. Appendix B presents additional visual results on diverse wild datasets to validate real-world robustness.

\vspace{0.5em}
\noindent\textbf{A. Theoretical Analysis of Data-Dependent Coupling}
\vspace{0.5em}

\noindent\textbf{Property 1 (Reduced Conditional Velocity Variance).}  \textit{Let $\bm{z}_t = (1-t)\bm{z}_0 + t\bm{z}_1$ be the probability path. The conditional variance of the velocity field $\lambda_t(\bm{z}) = \mathbb{E}[ \| (\bm{z}_1 - \bm{z}_0) - \bm{v}_t(\bm{z}) \|^2 \mid \bm{z}_t = \bm{z} ]$ is strictly reduced under our data-dependent coupling $\rho(\bm{z}_0, \bm{z}_1) = \rho_1(\bm{z}_1)\rho(\bm{z}_0 \mid \bm{z}_1)$ compared to the independent coupling $\rho_{\text{ind}}(\bm{z}_0, \bm{z}_1) = \rho_0(\bm{z}_0)\rho_1(\bm{z}_1)$.}

\vspace{0.5em}

\noindent\textbf{Proof.} From the linear interpolant $\bm{z}_t = (1-t)\bm{z}_0 + t\bm{z}_1$, the instantaneous velocity is 

\begin{equation}
    \dot{\bm{z}}_t = \bm{z}_1 - \bm{z}_0 = \frac{1}{t}(\bm{z}_t - \bm{z}_0).
\end{equation}
For a fixed state $\bm{z}_t = \bm{z}$ at time $t$, the conditional velocity variance $\lambda_t(\bm{z})$ is:
\begin{equation}
\label{eq:app_var_def}
\lambda_t(\bm{z}) = \mathrm{Var}(\dot{\bm{z}}_t \mid \bm{z}_t = \bm{z}) = \frac{1}{t^2} \mathrm{Var}(\bm{z}_0 \mid \bm{z}_t = \bm{z}).
\end{equation}

\textit{1) Independent Coupling Case:} Under independent coupling, $\bm{z}_0$ (typically $\mathcal{N}(\mathbf{0}, \mathbf{I})$) and $\bm{z}_1$ are sampled independently. The condition $\bm{z}_t = \bm{z}$ defines a broad posterior:

\begin{equation}
    \rho_{\text{ind}}(\bm{z}_0 \mid \bm{z}_t = \bm{z}) \propto \rho_0(\bm{z}_0) \rho_1\left(\frac{\bm{z} - (1-t)\bm{z}_0}{t}\right).
\end{equation}
Since $\rho_0$ and $\rho_1$ represent the entire source and target manifolds, for any given $\bm{z}$, there exists a vast set of $(\bm{z}_0, \bm{z}_1)$ pairs that can intersect at $\bm{z}$. This results in a high conditional variance $\mathrm{Var}_{\text{ind}}(\bm{z}_0 \mid \bm{z}_t = \bm{z}) \approx \text{Global Variance}$, which forces the marginal velocity $\bm{v}_t(\bm{z})$ to average over many conflicting directions, causing high trajectory curvature.

\textit{2) Data-Dependent Coupling Case:} In SCFlowFR, we model the ill-posed relationship as $\bm{z}_0 = \mathcal{G}(\bm{z}_1) + \bm{\eta}$, where $\mathcal{G}$ is the degradation process and $\bm{\eta} \sim \mathcal{N}(\mathbf{0}, \sigma_\eta^2 \mathbf{I})$ accounts for the intrinsic posterior uncertainty. Substituting the path constraint $\bm{z}_1 = \frac{1}{t}(\bm{z} - (1-t)\bm{z}_0)$ into this coupling yields an implicit constraint on $\bm{z}_0$:
\begin{equation}
\label{eq:app_implicit}
\bm{z}_0 = \mathcal{G}\left( \frac{\bm{z} - (1-t)\bm{z}_0}{t} \right) + \bm{\eta}.
\end{equation}

Assuming $\mathcal{G}$ is locally Lipschitz with constant $L < \infty   $ and the Jacobian is non-singular, by the implicit function theorem we have
\begin{equation}
    \mathrm{Var}_{\text{dep}}(\bm{z}_0 \mid \bm{z}_t = \bm{z}) \leq \frac{\sigma_\eta^2}{(1 - L \cdot t/(1-t))^2},
\end{equation}
which is $O(\sigma_\eta^2)$ and strictly smaller than the global variance when $\sigma_\eta^2 \ll \mathrm{Var}_{\text{ind}}$. This tight coupling reduces path crossovers, aligning $  \bm{v}_t(\bm{z})  $ with the linear trajectories $  \bm{z}_1 - \bm{z}_0  $ and linearizing the transport as $\sigma_\eta \to 0$.

\vspace{0.5em}
\noindent\textbf{Property 2 (Reduced Transport Kinetic Energy).}
\textit{For the probability flow ODE $\dot{\bm{Z}}_t = \bm{v}_t(\bm{Z}_t)$ where $\bm{Z}_0 \sim \rho_0$ and $\bm{Z}_1 \sim \rho_1$, the expected kinetic energy is bounded by the expected squared displacement of the coupling: $\mathbb{E} \int_0^1 \|\dot{\bm{Z}}_t\|^2 \, dt \leq \mathbb{E}[\|\bm{z}_1 - \bm{z}_0\|^2]$. Our data-dependent coupling minimizes this bound compared to independent coupling.}

\vspace{0.5em}

\noindent\textbf{Proof.} By the definition of the marginal velocity field in Eq.\,\eqref{eq:velocity_expectation}, $\bm{v}_t(\bm{z}) = \mathbb{E}[\bm{z}_1 - \bm{z}_0 \mid \bm{z}_t = \bm{z}]$. Applying Jensen's Inequality to the convex function   yields:
\begin{equation}
\|\bm{v}_t(\bm{z})\|^2 \leq \mathbb{E}[\|\bm{z}_1 - \bm{z}_0\|^2 \mid \bm{z}_t = \bm{z}].
\end{equation}
Taking the expectation over $\bm{z}_t \sim \rho_t(\bm{z})$ and applying the law of iterated expectations:
\begin{equation}
\mathbb{E}_{\rho_t}[\|\bm{v}_t(\bm{z}_t)\|^2] \leq \mathbb{E}_{\rho_t}[\mathbb{E}[\|\bm{z}_1 - \bm{z}_0\|^2 \mid \bm{z}_t]] = \mathbb{E}_{\rho}[\|\bm{z}_1 - \bm{z}_0\|^2].
\end{equation}
Integrating over $t \in [0, 1]$ and noting that the RHS is a constant determined by the coupling $\rho(\bm{z}_0, \bm{z}_1)$, we obtain:
\begin{equation}
\label{eq:app_energy_bound}
\mathbb{E} \int_0^1 \|\dot{\bm{Z}}_t\|^2 \, dt = \int_0^1 \mathbb{E}_{\rho_t}[\|\bm{v}_t(\bm{z}_t)\|^2] dt \leq \mathbb{E}_{\rho}[\|\bm{z}_1 - \bm{z}_0\|^2].
\end{equation}
The expected squared displacement $\mathbb{E}[\|\bm{z}_1 - \bm{z}_0\|^2]$ represents the transport cost between the source and target distributions. 

\textit{1) In the Independent Case}, where $\bm{z}_0$ is sampled regardless of $\bm{z}_1$, the expectation becomes $\mathbb{E}[\|\bm{z}_1\|^2] + \mathbb{E}[\|\bm{z}_0\|^2] - 2\mathbb{E}[\bm{z}_1]^\top \mathbb{E}[\bm{z}_0]$. Since $\bm{z}_0$ is typically a zero-mean Gaussian $\mathcal{N}(\mathbf{0}, \mathbf{I})$, this term is dominated by the global second moments of both manifolds, leading to a large transport cost and highly inefficient, long-range trajectories.

\textit{2) In our Data-Dependent Case}, we utilize the structural dependency between $\text{LQ}$ and $\text{HQ}$. Letting $\bm{z}_0 = \mathcal{E}(\text{LQ}) + \bm{\varepsilon}$, where $\mathcal{E}(\text{LQ})$ is semantically aligned with the target $\bm{z}_1$, the displacement becomes:
\begin{equation}
\mathbb{E}[\|\bm{z}_1 - (\mathcal{E}(\text{LQ}) + \bm{\varepsilon})\|^2] = \mathbb{E}[\|\bm{z}_1 - \mathcal{E}(\text{LQ})\|^2] + \sigma^2.
\end{equation}
Because $\mathcal{E}(\text{LQ})$ provides a coarse estimate of $\bm{z}_1$, the term $\|\bm{z}_1 - \mathcal{E}(\text{LQ})\|^2$ is restricted to the residual reconstruction error, which is significantly smaller than the global variance.

\vspace{0.5em}
\noindent\textbf{B. Additional Visual Results}
\vspace{0.5em}

\begin{figure}[ht]
  \centering
  \includegraphics[width=1\linewidth]{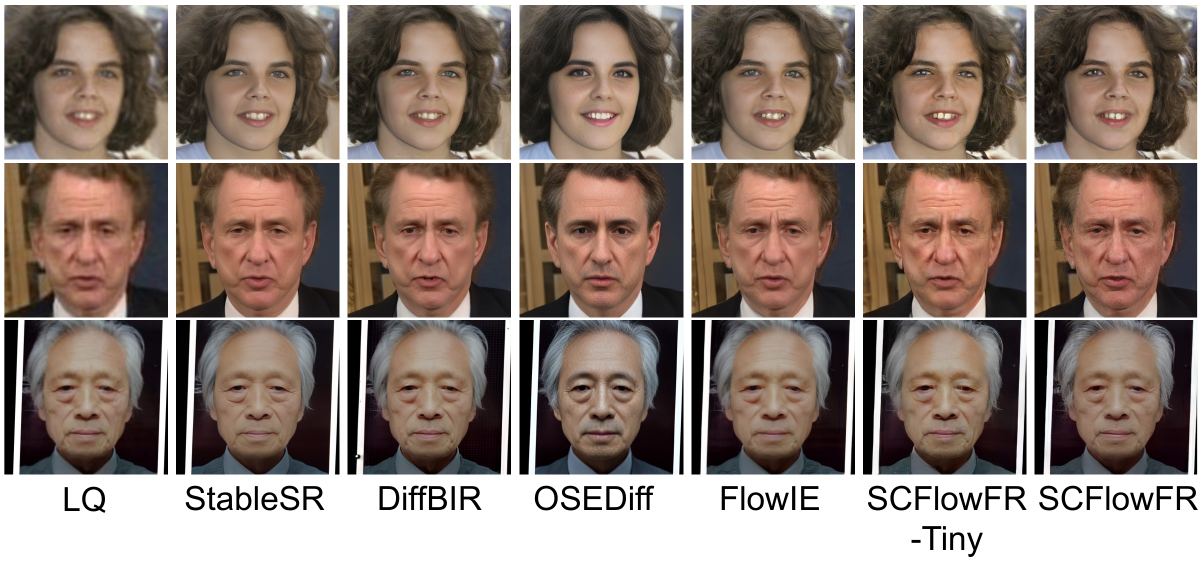}
  \caption{Qualitative comparisons on three wild datasets. From top to bottom: samples from the CelebChild, LFW, and WebPhoto datasets, respectively.}
  \label{fig:appendix_wild}
  \vspace{-1em}
\end{figure}

As illustrated in Fig.\,\ref{fig:appendix_wild}, our approach excels in restoring fine facial details, generating more realistic hair strands with clearer textures and preserving authentic skin texture with wrinkles. These visual improvements align with the quantitative gains, demonstrating the model's ability to produce perceptually convincing reconstructions in unconstrained wild scenarios.

\end{document}